# Classification of electromagnetic interference induced image noise in an analog video link

Anthony Purcell [1,2], Ciarán Eising [2]

[1] *Analog Devices, Inc*
[2] *Dept. Of Electronic and Computer Engineering, University of Limerick*

**Abstract**

With the ever-increasing electrification of the vehicle showing no sign of retreating, electronic systems deployed in automotive applications are subject to more stringent Electromagnetic Immunity compliance constraints than ever before, to ensure the proximity of nearby electronic systems will not affect their operation. The EMI compliance testing of an analog camera link requires video quality to be monitored and assessed to validate such compliance, which up to now, has been a manual task. Due to the nature of human interpretation, this is open to inconsistency. Here, we propose a solution using deep learning models that analyse, and grade video content derived from an EMI compliance test. These models are trained using a dataset built entirely from real test image data to ensure the accuracy of the resultant model(s) is maximised. Starting with the standard AlexNet, we propose four models to classify the EMI noise level.

**Keywords:** Machine Learning, Image Noise, EMI, SNR, Regularisation

## 1 Introduction

The problem that this research aims to address is one in the domain of automotive electronics, more specifically, the domain of automotive video electronics. The addition of robust video interfaces to vehicles is something that has grown considerably over the past decade, but the idea has been around for much longer than that. Ever since the first reversing camera was fitted to the Buick Centurion back in the 1950's, the addition of video and camera interfaces to vehicles has progressed considerably and become much more commonplace since then. Most vehicles being sold today are being presented with reversing camera options, with many manufacturers including them as standard. In 2018, all new cars sold in the US were mandated to have a reversing camera fitted as standard [US, Department of Transportation 2014], and in the EU, reversing cameras along with a host of other safety applications for cameras will become mandatory in 2022 [European Commission 2019]. The number of cameras that have been deployed in new vehicles has grown year on year over the past five years and this growth is predicted to continue [Analog Devices 2018]. These interfaces will enable both safety critical and non-safety critical applications. For example, cameras are set to replace external and internal mirrors, perform critical Advanced Driver-Assistance Systems (ADAS) functionality, such as collision avoidance and driver status monitoring, as well as enabling full surround-view monitoring of the vehicle, blind spot detection and night vision, to name but a few novel applications. This increase in both the number of cameras in the vehicle, coupled with the more increasing safety critical nature of the applications that they service, means that the task of system level design, validation, and Electromagnetic Compatibility (EMC) compliance verification that needs to be addressed as part of the vehicles design is more important than ever. This research aims to address the topic of EMC compliance, specifically, the challenges often encountered during the Electromagnetic Immunity (EMI) validation process of a video link destined for an automotive application, and how the application of machine learning can enable more consistent testing methodologies, yielding more meaningful and comparable test results between different systems and system configurations.

## 2   State of the Art

This section will frame the problem being overcome by this research. Image noise is something that can exist in many forms, so the type of noise being considered will be formally defined. As well as this, a widely used method for determining an images noise level will be described, and how ML techniques can overcome its shortcomings.

In a camera system, image noise can come from many sources. For example, image sensors are known to introduce Gaussian noise to a system, and image processing operations can introduce quantisation noise [Kleinmann and Wueller 2007]. The noise being considered here is noise that is delivered via an external source or interferer. That interferer is introduced in the form of a Bulk Current Injection (BCI) EMI test [Mahesh and Subbarao 2008]. The BCI test method is performed by using a current probe acting as a transformer to inject a current of a specified magnitude and frequency onto the cable harness at defined positions relative to the Device Under Test (DUT). It tests immunity performance in the frequency range of 1 MHz to 400 MHz [ISO 2011]. This noise is periodic in nature, is well defined and introduced in a controlled manner during an immunity test. It is noise of specifically this nature that will be the focus of discussion here. Figure 1 gives an example of noise injection on a test pattern image.

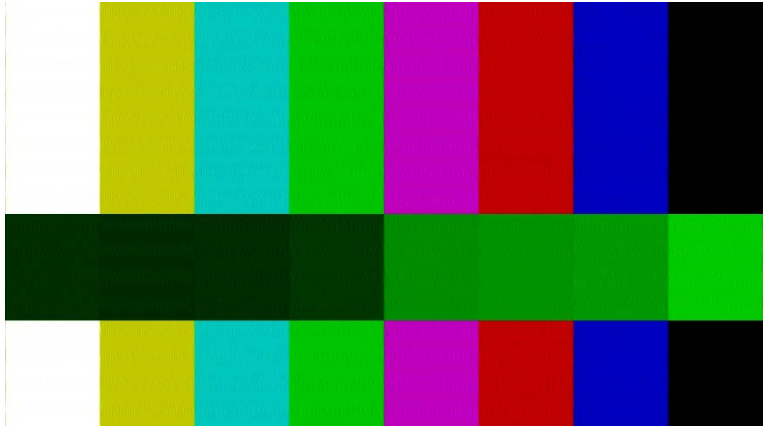

Figure 1: BCI Noise Injection on a Colour Bar Test Pattern

The classical approach to classifying noise in an image, given a reference image, is by using the peak signal-to-noise ratio (PSNR). For two Images $I_1$ and $I_2$ that have a 2-dimentional size of $i$ and $j$ and are composed of $c$ channels, the PSNR in dB can be defined as:

$$PSNR = 10 \cdot log_{10}\left(\frac{MAX_I^2}{\frac{1}{c*i*j}\Sigma(I_1 - I_2)^2}\right) dB$$

Where $MAX_I$ is largest valid pixel value in the Images $I_1$ and $I_2$, usually 255 for RGB images [OpenCV 2019]. If the images are identical, the divisor will be 0, and any difference in the images will yield a finite value for the image noise present. There are some short comings of the SNR method outlined above for the application that is the subject of this paper.

One advantage of using ML techniques to address the need here is that it will allow specific, visual noise to be targeted as the criteria for classifying the image quality. SNR quantification will highlight any variance between the pixel values of two images. Negligible changes in pixel values will lead to higher PSNR for, even though these differences may not be visible to the human eye. In effect, it has the potential to flag image variance that is not of interest to an inherently qualitative test, where visible noise is the main criteria to judge a pass/failure. Furthermore, the focus of this system will be on an analog video link. Analog video links are inherently not bit-accurate links, where from frame to frame, identical image content under perfect operating conditions is likely to show some least significant bit (LSB) variance. This variance would be considered as noise by SNR analysis, even if it is only ever likely to be a very low value. Thus, we can say, it is desirable to have a machine learning algorithm that models subjective noise analysis. Others have investigated ways through which this gap between measurement and physical interpretation can be bridged. Kleinmann and Wueller have evaluated methods that focus on the difference between quantifying image noise based on the perception of the human eye as opposed to purely a measurement focussed approach. They investigated two algorithms that aim to emulate the process of the human visual system more closely than the SNR measurement approach:

1. A model for visual noise measurement where the process of human vision is simulated using opponent colour space and contrast sensitivity functions to come up with a visual noise value [Hung, Enomoto and Kozo 1996].
2. The S-CIELab model, which simulates human vision in approximately the same way as the model by Hung, et al, with the addition of an image comparison using the CIEDE2000 colour difference formula, which was designed for predicting the visual perceived difference between colour images [Fairchild and Johnson 2003].

Both models yielded good results, highlighting differences between the simulated visual noise value and the SNR measured value. In saying this, both models showed restrictions during their respective evaluations. The former model is effective at evaluating noise for uniform colour patches, so this places a restriction on the image content to be used in the testing. The performance of the latter model was deemed to be influenced by the kind of noise seen in the image, so is not expected to operate consistently across all image noise types. This would mean that it would need to be thoroughly evaluated to be used and trusted for a specific application [Kleinmann and Wueller 2007].

## 3 The Dataset

Fundamental to the effectiveness of any machine learning approach to solving a problem is an appropriate dataset. There was not an appropriate pre-existing dataset available to utilise for this work, so the dataset that will be used was created from scratch from video frames captured from a real test environment described in Figure 2.

### 3.1 Data capture

A common method for classifying image quality in an EMC environment involves capturing and transmitting the video that is sent over the camera link out of the test chamber via a HDMI optical extender. This system is a purpose-built system to ensure EMI does not affect the received video that is being sent out of the chamber for viewing. Outside the chamber, a HDMI monitor can be used to capture the video for live and post analysis. The setup is demonstrated in Figure 2. For a general introduction to BCI test methods, the reader is referred to [Mahesh and Subbarao 2008].

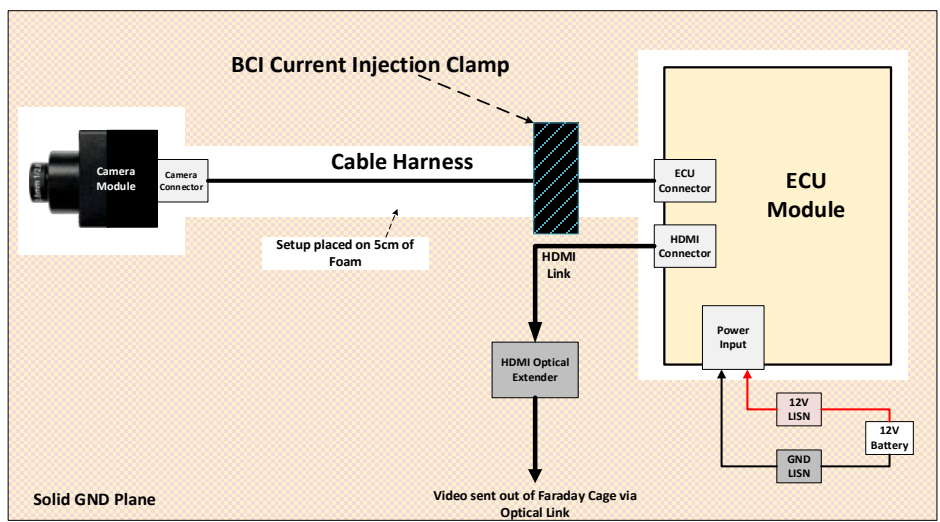

Figure 2: Sample BCI Test Setup for a Single Camera Link

## 3.2 The Video Content

The content of the video data being transmitted over the link is also an important factor to be considered in the creation of the dataset. Real-world image sensor data captured in an EMI chamber can prove difficult to execute consistent video quality analysis on, because the image is usually quite dark (or at least certainly darker than would be expected during normal operation), which can make the noise and the genuine image content difficult to distinguish. For this reason, video test patterns are used for image noise analysis. This approach comes with three significant benefits:

1. The image content of interest is consistent and does not change from a visual perspective from frame to frame. This makes identifying and analysing induced noise an easier task, and the test a more stringent one, because noise seen here may well be indistinguishable on a small in-vehicle display showing real world content.
2. The artificial nature of the test pattern content allows a wide spectrum of colour content to be examined, meaning a more thorough analysis can be done across a wider brightness and colour spectrum than would be possible with a natural image.
3. Finally, we are interested only in the noise introduced during the analog video transmission. The transmission of the test pattern isolates this source of noise from any other noise source.

A colour bar test pattern is selected as this contains low frequency video content which makes identifying external interference a less strenuous visual task.

## 3.3 The Dataset Structure

The dataset is formulated into training, validation, and test sets. We define the categories that the dataset will assume when communicating noise interference:
- Level 1: No visual noise detected
- Level 2: Low level of visual noise detected
- Level 3: Medium level of noise detected
- Level 4: High level of noise detected
- Level 5: Video link has ceased to operate, loss of lock event, flat blue field displayed.

These categories are, of course, arbitrary, but due to the subjective nature of the task, this is somewhat unavoidable. Care was taken to ensure consistency across all samples within a category, but this cannot be guaranteed from merely a visual segregation process. The dataset is organised into 800, 200 and 100 image samples within each category of the training, validation, and test sets respectively. Figure 3 shows examples of noise levels 2 to 4 (level 5 is not shown, as this is simply a flat blue field).

# 4 Video Noise Grading Model

This section will describe the implementation and verification of candidate CNN models to undertake the task of performing a video noise classification task. The creation of a custom dataset as described in section 3 will be utilised to train and validate these CNN's. For reasons that will be discussed in detail, four candidate models are proposed and their difference in terms of complexity and performance is discussed comparatively. In addition to this, particular focus is given to techniques that result in improved model accuracy and generalisation.

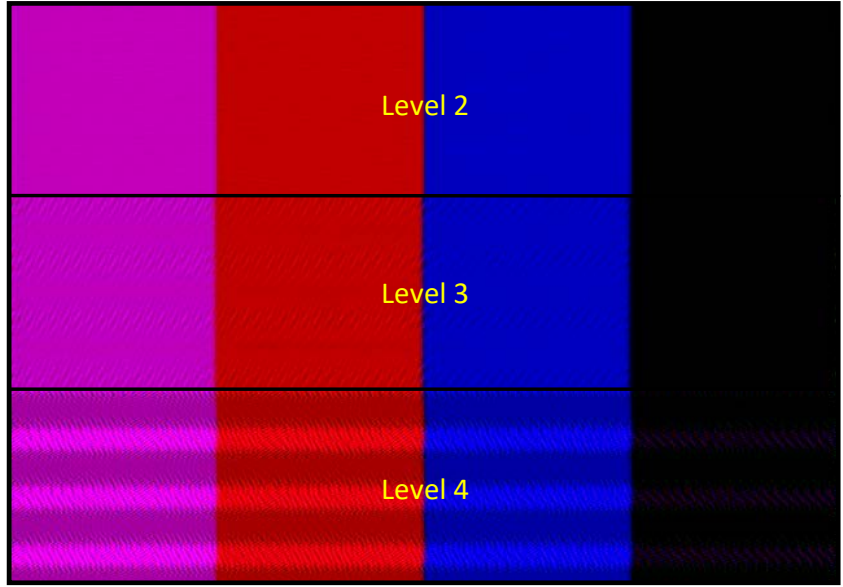
Figure 3: Example of the Interclass Image Quality Variance on a patch of the test pattern.

## 4.1 Data Pre-processing

The dataset is comprised of frames of 1280x720p video in the YCbCr colour space. These frames are reduced in size using nearest neighbour interpolation before being input to the network. Whilst a network that could take these images in their default state is certainly possible, it was advantageous from a complexity and training time perspective to reduce the size of the images appropriately before applying them to the network. A concern from performing such an operation is that once the images are resized, the features that are required to be detected (the image noise, in this case) should not be lost or degraded significantly in the process, which may prevent a model's ability to detect the feature. The AlexNet CNN architecture was used as a starting point for the model development. This architecture mandates a 227x227 size input, so the images were resized to meet this requirement. Across all models that are proposed, this input size remained unaltered for comparison. Rescaling of the image pixel values was also applied. By default, a pixel can occupy any value from 0 to 255 for 8-bit images. These pixel values are rescaled from [0…255] to [0… 1]. The final alteration that was made to the input dataset before being applied to a model was that the images were converted to greyscale, dropping the chroma channels of the image (Figure 4). This allows for the discarding of two feature maps in the input to the network, thus reducing the complexity of the required network. The encoding scheme on the video link used here ensures the luma and chroma portions of the signal are equally susceptible to EMI interference, thus using just the luma data is sufficient to characterise the noise level.

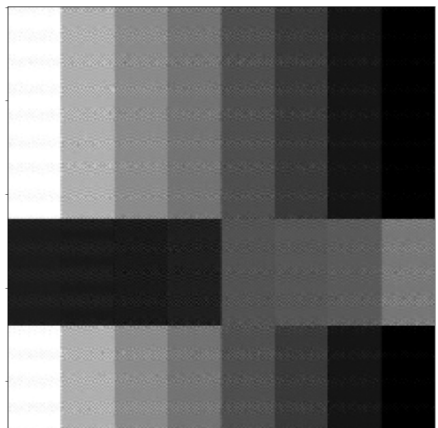
Figure 4: Greyscale Image Showing the Retention of Noise Features

## 4.2 Model Definition and Structure

Four CNN models will be defined for to carry out the task of classifying image quality. The models presented here start with the well-known AlexNet structure [Krizhevsky, Sutskever and Hinton 2012], and subsequent models with lower complexity are then defined. To start with, the AlexNet model was left as standard, apart from the following details:

1. The input size of the network. Originally, this model was applied to RGB images (227x227x3), but here the input size is reduced to 227x227x1 due to the conversion to greyscale applied to the images.
2. Data augmentation is applied to the input images before the convolutional layers. The images were randomly flipped in both the horizontal and vertical direction, which ensures the noise profile is maintained, but the active image content is differed, encouraging the models to better characterise the noise interference.

### 4.2.1 Lower Complexity CNN Models

For comparison with the AlexNet implementation, models that represent significant reductions in the number of trainable parameters are proposed to see if this task can be achieved more effectively using a less complex network. These networks are built from the same building blocks that constructed the AlexNet model, and use the same data augmentation step at the beginning. The main differences are in the convolution layers (both in the parameters specified and the number of layers) and in the number and size of the fully connected layers of the networks. A comparison of the three lower complexity models' structure is given in Figure 5. All activations were ReLU apart from the final fully connected layer in each model, which has a softmax activation.

| CNN Model 2 | | CNN Model 3 | | CNN Model 4 | |
|---|---|---|---|---|---|
| Input | 227x227x1 | Input | 227x227x1 | Input | 227x227x1 |
| Conv2D | 55x55x32 | Conv2D | 55x55x32 | Conv2D | 55x55x16 |
| Activation | 55x55x32 | Activation | 55x55x32 | Activation | 55x55x16 |
| Max Pool | 27x27x32 | Max Pool | 27x27x32 | Max Pool | 27x27x16 |
| Conv2D | 27x27x96 | Conv2D | 27x27x64 | Conv2D | 27x27x16 |
| Activation | 27x27x96 | Activation | 27x27x64 | Activation | 27x27x16 |
| Max Pool | 13x13x96 | Max Pool | 13x13x64 | Max Pool | 13x13x16 |
| Conv2D | 13x13x128 | Conv2D | 13x13x128 | Flatten | 2704 |
| Activation | 13x13x128 | Activation | 13x13x128 | FC | 50 |
| Max Pool | 6x6x128 | Max Pool | 6x6x128 | FC | 5 |
| Flatten | 4608 | Flatten | 4608 | | |
| FC | 512 | FC | 100 | | |
| FC | 5 | FC | 5 | | |
| Total Params: | 2,504,741 | Total Params: | 557,661 | Total Params: | 137,777 |

Figure 5: Lower Complexity CNN Model Structure

## 5 Results

The training of each model was done with the Adam optimiser with a learning rate of $10^{-3}$. All models converged within 30 epochs, and this number of training cycles was kept standard across all models for comparison. The performance of this training process is captured below in Table 1 and Figure 6.

| | AlexNet Model | | | CNN Model 2 | | | CNN Model 3 | | | CNN Model 4 | | |
|---|---|---|---|---|---|---|---|---|---|---|---|---|
| Category | Precision | Recall | F1-Score | Precision | Recall | F1-Score | Precision | Recall | F1-Score | Precision | Recall | F1-Score |
| Level 1 | 1.00 | 1.00 | 1.00 | 1.00 | 1.00 | 1.00 | 1.00 | 1.00 | 1.00 | 1.00 | 1.00 | 1.00 |
| Level 2 | 1.00 | 0.91 | 0.95 | 1.00 | 0.69 | 0.82 | 1.00 | 0.76 | 0.86 | 1.00 | 1.00 | 1.00 |
| Level 3 | 0.64 | 1.00 | 0.78 | 0.97 | 1.00 | 0.99 | 0.96 | 1.00 | 0.98 | 0.94 | 1.00 | 0.97 |
| Level 4 | 0.83 | 0.44 | 0.58 | 0.78 | 1.00 | 0.88 | 0.80 | 0.96 | 0.87 | 1.00 | 0.94 | 0.97 |
| Level 5 | 1.00 | 1.00 | 1.00 | 1.00 | 1.00 | 1.00 | 1.00 | 1.00 | 1.00 | 1.00 | 1.00 | 1.00 |
| Accuracy | | | 0.87 | | | 0.94 | | | 0.94 | | | 0.99 |

Table 1: Performance Metrics on Test Set Data

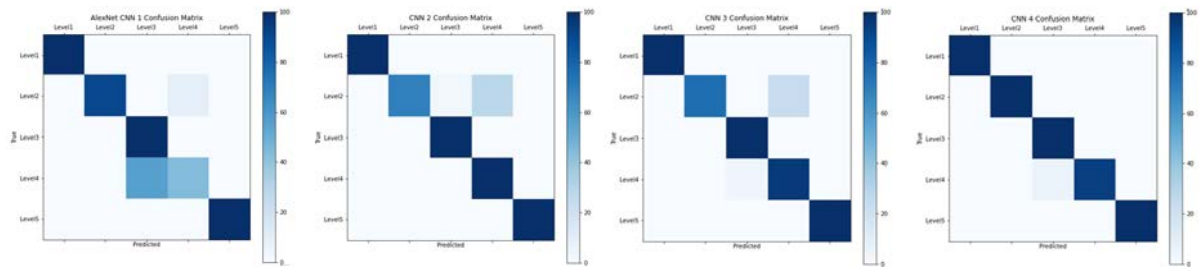

Figure 6: Confusion Matrix on Test Set Data

The metrics and confusion matrix comparisons show the lower complexity models generalise better, giving the best out of sample performance. We can see the issues experienced by model 1 (AlexNet) in classifying level 4 samples, clearly showing the reported recall value of < 50%. Models 2 and 3 are showing issues misclassifying some level 2 samples. These issues are resolved in the lowest complexity model, where near perfect performance is shown. The reduction in model complexity from the AlexNet model was necessary to achieve the best performance, and based on the complete analysis of model performance, model 4 would be the best model to use. This can be attributed to the regularising effect of reducing the number of parameters the model has to fit the data. This, however, does not mean that the AlexNet model cannot be made to work effectively for this task. To explore this further, the next section will discuss using explicit regularisation techniques to improve performance.

### 5.1 Regularisation

The more complex a model is, the more likely overfitting to the training data is to occur. To test if this is causing the more complex models' poorer out of sample performance, regularisation is applied. L2 regularisation is applied with a $\lambda$ value of 0.01. Once this addition was made, all 4 models were re-trained. The performance metrics, for comparison, are shown below in Table 2 and Figure 7.

|          | AlexNet Model - L2 Reg | | | CNN Model 2 - L2 Reg | | | CNN Model 3 - L2 Reg | | | CNN Model 4 - L2 Reg | | |
|---|---|---|---|---|---|---|---|---|---|---|---|---|
| Category | Precision | Recall | F1-Score | Precision | Recall | F1-Score | Precision | Recall | F1-Score | Precision | Recall | F1-Score |
| Level 1 | 1.00 | 1.00 | 1.00 | 1.00 | 1.00 | 1.00 | 1.00 | 1.00 | 1.00 | 1.00 | 1.00 | 1.00 |
| Level 2 | 0.95 | 0.93 | 0.94 | 1.00 | 0.91 | 0.95 | 0.93 | 0.95 | 0.94 | 1.00 | 0.78 | 0.88 |
| Level 3 | 1.00 | 0.94 | 0.97 | 0.99 | 0.99 | 0.99 | 0.62 | 1.00 | 0.77 | 0.76 | 1.00 | 0.87 |
| Level 4 | 0.88 | 0.95 | 0.91 | 0.91 | 0.99 | 0.95 | 1.00 | 0.37 | 0.54 | 1.00 | 0.91 | 0.95 |
| Level 5 | 1.00 | 1.00 | 1.00 | 1.00 | 1.00 | 1.00 | 1.00 | 1.00 | 1.00 | 1.00 | 1.00 | 1.00 |
| Accuracy |  |  | 0.96 |  |  | 0.98 |  |  | 0.86 |  |  | 0.94 |

Table 2: Performance Metrics on Test Set Data with L2 Regularisation

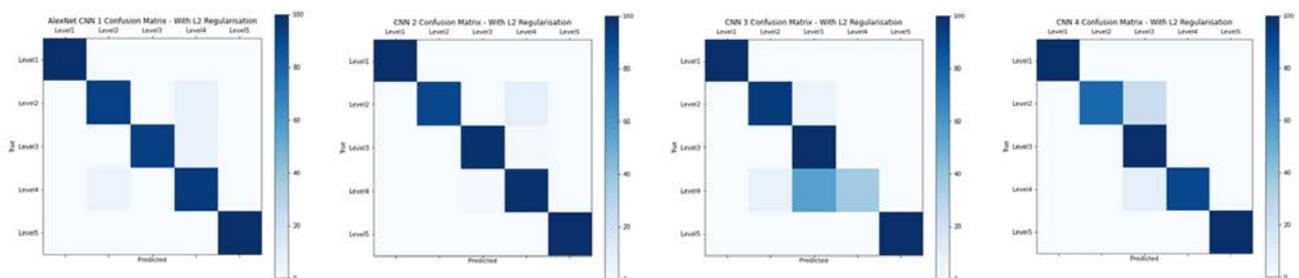

Figure 7: Confusion Matrix on Test Set Data with L2 Regularisation

From the above results, models 1 and 2 benefitted from the addition of L2 regularisation, with model 2 being the best performing of the set. Regularisation reduced overfitting such that out of sample performance is significantly improved, with an increase of 9% in accuracy in the case of the AlexNet model. Models 3 and 4 showed poorer out of sample performance because of the regularisation, the amount of regularisation applied ($\lambda = 0.01$) is clearly causing significant bias. Reducing this $\lambda$ value would remedy this induced bias in these models.

# 6   Conclusions

This paper has shown that the task of assigning a class that represents image noise content is not only achievable but can be done quite effectively using ML techniques. Up until now, this has been a manual task for technicians. Furthermore, it is shown that all models that were initially presented as candidates can be made to perform better through some considered techniques to alter and constrain the training process to ensure an appropriate level of overfitting of the training data is realised. These models represent a wide range of complexity levels, the more complex models showing as good candidates for further development if the dataset to hand increases in size or variance. The exercise here has also proven that the task of classifying video quality in this fashion can be achieved and implemented on a wide range of hardware capabilities, from cost effective SoC's to large GPU devices.